%% file: bmvc_final.tex
\documentclass{bmvc2k}
\usepackage{amsfonts,amssymb}
\usepackage{booktabs}
\usepackage{upgreek}
\usepackage{enumitem}
\usepackage{bm}
\usepackage{censor}

\definecolor{Clean}{RGB}{52,194,48}
\definecolor{RedNew}{RGB}{242, 133, 133}
\definecolor{NewSky}{RGB}{89, 213, 224}

\usepackage{multirow}
\usepackage{makecell}
\usepackage{subfigure}

\usepackage{xspace}
\usepackage{interval}
\usepackage{bm,upgreek}
\usepackage{colortbl}
\usepackage{enumitem}
\usepackage{amssymb}
\usepackage{pifont}
\usepackage{graphicx}
\usepackage{color}
\usepackage{multirow}
\usepackage{amsmath,amssymb}
\usepackage{mathtools,leftindex,tensor,mhchem}

\usepackage{pifont}
\usepackage{graphics,color}
\usepackage[dvipsnames]{xcolor}
\usepackage{bm}
\usepackage{textcomp}

\DeclareMathOperator*{\argmax}{arg\,max}
\DeclareMathOperator*{\argmin}{arg\,min}

\usepackage{printlen}

\title{Calibration-Aware Prompt Learning for Medical Vision-Language Models}

\addauthor{Abhishek Basu}{abhishek.basu@mbzuai.ac.ae}{1}
\addauthor{Fahad Shamshad}{fahad.shamshad@mbzuai.ac.ae}{1}
\addauthor{Ashshak Sharifdeen}{ashshak.sharifdeen@mbzuai.ac.ae}{1}
\addauthor{Karthik Nandakumar}{nandakum@msu.edu}{1,2}
\addauthor{Muhammad Haris Khan}{muhammad.haris@mbzuai.ac.ae}{1}

\addinstitution{
Department of Computer Vision,    \\
Mohamed bin Zayed University of
Artificial Intelligence (MBZUAI), \\
Abu Dhabi, UAE
}
\addinstitution{
 Department of Computer Science and Engineering, \\
 Michigan State University (MSU),\\
 East Lansing, USA
}

\runninghead{BASU ET.AL}{CalibPrompt}

\begin{document}

\maketitle

\begin{abstract}
Medical Vision-Language Models (Med-VLMs) have demonstrated remarkable performance across diverse medical imaging tasks by leveraging large-scale image-text pretraining. However, their confidence calibration is largely unexplored, and so remains a significant challenge. As such, miscalibrated predictions can lead to overconfident errors, undermining clinical trust and decision-making reliability.
To address this, we introduce \texttt{CalibPrompt}, the first framework to calibrate Med-VLMs during prompt tuning. 
\texttt{CalibPrompt} optimizes a small set of learnable prompts with carefully designed calibration objectives under scarce labeled data regime.
First, we study a regularizer that attempts to align the smoothed accuracy with the predicted model confidences. Second, we introduce an angular separation loss to maximize textual feature proximity toward improving the reliability in confidence estimates of multimodal Med-VLMs.
Extensive experiments on four publicly available Med-VLMs and five diverse medical imaging datasets reveal that \texttt{CalibPrompt} consistently improves calibration without drastically affecting clean accuracy. Our code is available at \url{https://github.com/iabh1shekbasu/CalibPrompt}. 
\end{abstract}

\section{Introduction}
\label{sec:intro}
Medical Vision-Language models (Med-VLMs) have emerged as powerful tools for medical image analysis, leveraging large-scale image-text pretraining to enable zero-shot classification across diverse medical imaging tasks~\cite{moor2023foundation,chia2024foundation,zhao2023clip}. These models align medical images with textual descriptions, facilitating interpretation and diagnosis without requiring task-specific fine-tuning. 
However, despite their strong performance in recognizing medical concepts, Med-VLMs often suffer from poor calibration, where their confidence scores fail to reliably indicate actual correctness~\cite{huang2020tutorial,nori2023can}, which is particularly concerning in medical imaging, as miscalibrated model can lead to misdiagnoses and undermine clinical trust~\cite{lambert2024trustworthy}.

Model calibration techniques generally fall into two categories: post-hoc calibration and training-time calibration. Post-hoc methods, such as Platt scaling~\cite{platt1999probabilistic} and temperature scaling~\cite{guo2017calibration}, adjust confidence scores after training via a transformation function. While computationally inexpensive, they have two key limitations: (1) reliance on a small validation set, which may not reflect real-world medical distributions~\cite{niculescu2005predicting,wang2024open}, and (2) failure to improve the model’s internal representations, leaving calibration issues unresolved at the decision-making level~\cite{ovadia2019can}. In contrast, training-time calibration jointly optimizes accuracy and calibration, leading to more robust and generalizable confidence estimates~\cite{kumar2018trainable}. By integrating calibration objectives into training, it ensures well-calibrated Med-VLM predictions across medical tasks, enhancing clinical trust. However, fine-tuning large-scale Med-VLMs with calibration objectives is often impractical due to high computational cost and requirement of massive labeled medical datasets.

Meanwhile, prompt tuning has emerged as an efficient alternative to full-model fine-tuning for adapting Med-VLMs to downstream tasks with limited data~\cite{zhou2022learning}. Unlike conventional fine-tuning, which updates the entire model, prompt tuning modifies only a small set of learnable parameters, significantly reducing computational costs while preserving generalization~\cite{hussein2024promptsmooth,hanif2024baple}. This efficiency is particularly valuable in medical imaging, where labeled data is scarce and full fine-tuning is often impractical. Despite its strong performance in data-limited settings, prompt tuning primarily optimizes classification and does not inherently improve model calibration. This raises a key question: \textit{Can the efficiency of prompt tuning, with its low data requirements and minimal parameter updates, be leveraged to enhance calibration without compromising adaptability?} Addressing this is crucial to ensure Med-VLMs produce both accurate and well-calibrated predictions for reliable clinical decision-making.

In this paper, we introduce \texttt{CalibPrompt}, the first approach to calibrate Medical Vision-Language Models during prompt learning. 
Specifically, we make following technical contributions. 
\begin{itemize}
    \item We investigate a simple regularizer that aligns softened accuracy with model confidences to effectively calibrate under class ambiguities inherent in medical imaging.

    \item We propose a novel angular separation loss that promotes angular gap between \textit{textual features} during prompt tuning, specifically tailored to the multimodal architecture of Med-VLMs.

    \item We demonstrate the effectiveness of \texttt{CalibPrompt} through comprehensive experiments across four publicly available Med-VLMs and five downstream datasets spanning different imaging modalities, achieving superior calibration performance while tuning only 0.1\% of the model parameters.
\end{itemize}


\section{Related Work}
\textbf{Medical Vision-Language Models (Med-VLMs)}. Med-VLMs inspired by Contrastive Language-Image Pretraining (CLIP), have advanced medical imaging by aligning image-text pairs across modalities such as X-ray, histopathology, and retinal imaging. These models enable zero-shot and few-shot classification, making them particularly valuable in data-scarce medical applications~\cite{moor2023foundation,chia2024foundation,zhao2023clip}. To adapt Med-VLMs efficiently, prompt learning (PL) has emerged as a lightweight alternative to full-model fine-tuning~\cite{zhou2022learning,hussein2024promptsmooth,hanif2024baple}. By introducing learnable prompt tokens without modifying the backbone, PL enhances task performance while maintaining computational efficiency, making it well-suited for medical imaging with limited data. In this work, we investigate whether PL can simultaneously improve calibration and task adaptation, positioning it as a parameter-efficient alternative to computationally intensive calibration methods for accurate and trustworthy predictions in medical AI.

\noindent \textbf{Confidence Calibration}. Confidence calibration assesses how well a model's predicted confidence aligns with its actual accuracy, a critical requirement in high-stakes domains like medical imaging. Well-calibrated models yield reliable uncertainty estimates, crucial for clinical decision-making, as overconfident misclassifications can lead to severe consequences~\cite{guo2017calibration,lambert2024trustworthy}. Post-hoc calibration methods, such as Temperature Scaling, adjust model logits via a learned temperature parameter optimized on a held-out validation set~\cite{guo2017calibration}. While computationally efficient, these methods heavily depend on labeled datasets closely matching the target distribution~\cite{niculescu2005predicting,wang2024open}.To overcome such limitations, train-time calibration methods incorporate calibration objectives directly into model training, typically through auxiliary loss functions alongside primary task-specific objectives, resulting in more robust and generalizable confidence estimates~\cite{kumar2018trainable,ovadia2019can,mdca}. For instance, MACSO~\cite{kugathasan2024matching} aligns predicted confidences with softened target distributions derived from the model's internal knowledge, utilizing correlation-based distance measures. Other effective train-time approaches include Margin-based Label Smoothing (MbLS)~~\cite{liu2022devil}, which imposes inequality constraints on logit distances to prevent overly confident predictions, and Logit Normalization (LogitNorm)~~\cite{wei2022mitigating}, which enforces a constant norm on logits during training to mitigate overconfidence. Moreover, Murugesan et al.~~\cite{murugesan2024robust} identified expanded logit distributions in prompt-tuned models as a significant calibration issue, introducing Zero-Shot Normalization to restore alignment with pretrained distributions. Test-time calibration methods like C-TPT~\cite{yoon2024c} addresses calibration via test-time prompt tuning by optimizing text feature dispersion using prototypes. Concurrent with our research, O-TPT~\cite{sharifdeen2025tpt} addresses calibration in vision-language models (VLMs) through test-time prompt tuning, enforcing strict orthogonality constraints on textual features without relying on labeled data. Similarly, Wang et al.~~\cite{wang2024open} proposed DAC, which adjusts softmax temperatures based on semantic distances between embeddings, primarily targeting novel-class calibration in domains with numerous classes. However, medical imaging datasets typically involve fewer classes, limiting DAC's applicability in specialized medical contexts. Their findings underscore that post-hoc calibration methods alone cannot fully recover the pretrained calibration behaviour of VLMs. In contrast, we introduce a novel prompt-based calibration framework that operates effectively in few-shot scenarios by jointly optimizing regularization objectives in both probability and feature spaces. Unlike O-TPT, our method encourages (rather than strictly enforces) angular separation, providing greater flexibility to capture nuanced class relationships within specialized medical imaging domains.

\section{Method}

Our goal is to calibrate Med-VLMs in data-limited settings to ensure that their confidence scores accurately reflect prediction correctness. In medical imaging, miscalibrated models can lead to overconfident errors with serious clinical implications. To this end, we introduce a new approach  \texttt{CalibPrompt} under prompt-learning setup which is built on two novel  regularizers. The first regularizer matches the softened accuracy with predicted confidences, while the second is an angular separation loss that explicitly maximizes the proximity between textual features. Below, we first describe zero-shot inference with Med-VLMs, then explain the prompt learning basics, and finally introduce our calibration-aware prompt tuning approach \texttt{CalibPrompt}.

\subsection{Preliminaries:}
\label{subsection:Preliminaries}
\textbf{Zero-Shot Inference for Med-VLMs:} Med-VLMs learn a joint representation of images and text through contrastive pretraining, enabling zero-shot classification. These models consist of an image encoder $\mathbf{E}_{\text{img}}: \mathcal{I} \rightarrow \mathbb{R}^d$ and a text encoder $\mathbf{E}_{\text{txt}}: \mathcal{T} \rightarrow \mathbb{R}^d$, where $\mathcal{I}$ and $\mathcal{T}$ denote the image and text spaces, respectively. Given an input image $\mathbf{I} \in \mathcal{I} \subseteq \mathbb{R}^{H \times W \times C}$, the image encoder extracts a $d$-dimensional feature vector $\mathbf{v} = \mathbf{E}_{\text{img}}(\mathbf{I})$. Similarly, the text encoder maps a textual prompt $\mathbf{t}(y) \in \mathcal{T}$ associated with class label $y \in \mathcal{Y}$ into a text feature vector $\mathbf{u} = \mathbf{E}_{\text{txt}}(\mathbf{t}(y))$. During zero-shot inference, class labels $\{y_1, \dots, y_K\}$ are converted into text prompts using a predefined template, such as $\mathbf{t}(y_i) = ``\texttt{A H\&E image of [CLASS} ~y_i\texttt{]}''$, and are processed by the text encoder to obtain $\{\mathbf{u}_1, \dots, \mathbf{u}_K\}$, where $\mathbf{u}_i = \mathbf{E}_{\text{txt}}(\mathbf{t}(y_i))$. 
For a test image $\mathbf{I}_t$, the similarity between the image and text features is computed as $s_i = \text{sim}(\mathbf{v}_t, \mathbf{u}_i)$, where $\mathbf{v}_t = \mathbf{E}_{\text{img}}(\mathbf{I}_t)$. The final classification probabilities are obtained using a softmax function as $\mathbb{P}(y_i | \mathbf{I}_t) = \frac{\exp(\tau s_i)}{\sum_{j=1}^{K} \exp(\tau s_j)}$,
where $\tau$ is the softmax temperature parameter. The predicted label is then given by $\hat{y}_t = \argmax_{y \in \mathcal{Y}} \mathbb{P}(y | \mathbf{I}_t)$.
The corresponding confidence is given by $\hat{p}_t = \max_{y \in \mathcal{Y}} \mathbb{P}(y | \mathbf{I}_t)$. While this zero-shot framework enables flexible classification, Med-VLMs often produce overconfident predictions as shown in Table~\ref{tab:zeroshot_results}. A naive solution is to fine-tune
Med-VLMs with explicit calibration objectives (i.e. train-time auxiliary losses); however, this is computationally
expensive and requires extensive labeled data. Thus, an efficient alternative is needed to enhance calibration without full model retraining.

\noindent \textbf{Prompt Learning:} Prompt learning (PL) has emerged as an efficient alternative, enabling adaptation to new tasks without modifying the model backbone. Instead of updating the entire network, PL optimizes a small set of learnable prompt tokens, making it particularly useful for data-scarce medical applications. When a text prompt $\mathbf{t}(y_i) \in \mathcal{T}$ is passed to the text encoder, it is tokenized into a sequence of word embeddings. Typically, a class-specific text prompt is represented as $[\mathbf{w}]_1~[\mathbf{w}]_2~\cdots~[\texttt{CLASS}~y_i]$, where each $[*]$ denotes a word embedding. In PL, all fixed embeddings (except for the class token) are replaced with $M$ learnable embeddings, transforming the prompt into $\mathbf{p}_{i1}~\mathbf{p}_{i2}~\cdots\mathbf{p}_{iM}~[\texttt{CLASS}~y_i]$, where each prompt embedding $\mathbf{p}$ has the same dimensionality as $[\mathbf{w}]$. Let $\mathcal{P} = {\mathbf{p}_{im}}$, where $i \in [1,K]$ and $m \in [1,M]$, represent the set of all learnable prompts. The output text feature vector, incorporating these learned prompts, is denoted as $\mathbf{u}_{i}(\mathcal{P})$, and the modified zero-shot classifier is $f_{\mathcal{P}}$.

\begin{figure}[t]
    \centering
    \label{fig:three_images}
    \includegraphics[width=\linewidth,trim={0 0.0cm 0 1cm},clip]{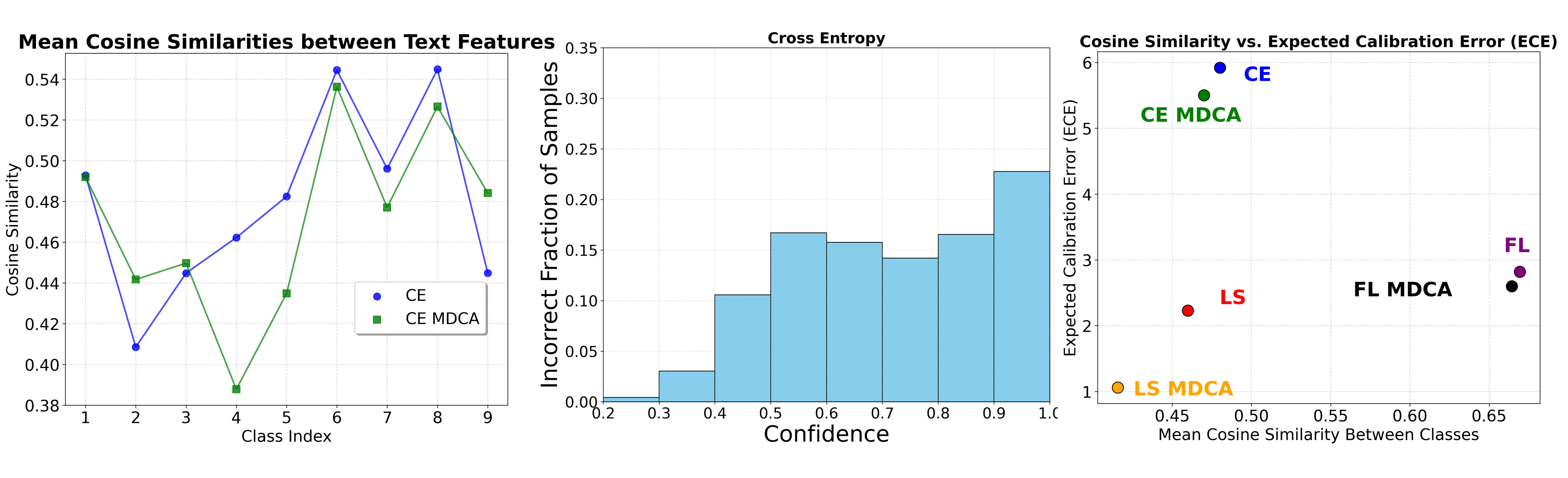}
    \caption{\textbf{Analysis of Prompt Learning Effects on Model Calibration}. \textit{Left}: Cross entropy (CE) shows higher text feature similarity between classes than CE MDCA. \textit{Middle}: Cross entropy histogram demonstrating overconfident misclassifications with higher confidence levels. \textit{Right}: Greater feature similarity (CE, CE MDCA) directly correlates with increased calibration error compared to regularized approaches (LS, FL MDCA).}
    \label{fig:fracCE+SLMDCA+angle}
\end{figure}

\noindent\textbf{Limitation and Motivation:} While prompt learning effectively adapts Med-VLMs to downstream tasks with limited data, our empirical analysis reveals a critical limitation for Med-VLMs: \textit{it increases calibration error despite improving classification accuracy}. We observe that prompt-tuned models consistently exhibit high Expected Calibration Error~~\cite{ECE}, indicating a mismatch between confidence scores and actual correctness.
To understand this behavior, we analyze the geometric properties of learned textual prompts and find that prompt tuning significantly increases intra-class cosine similarity, causing class representations to become highly aligned (see Fig.~\ref{fig:fracCE+SLMDCA+angle} left). While this enhances classification separability, it also amplifies confidence scores, leading to overconfident predictions and calibration error, with approximately 22\% of misclassifications occurring at high confidence levels of 0.9-1.0 (see Fig.\ref{fig:fracCE+SLMDCA+angle} middle). Further, our results reveal a strong correlation between high cosine similarity and increased miscalibration (see Fig.~\ref{fig:fracCE+SLMDCA+angle} right), underscoring the need for explicit text feature space regularization.

\subsection{\texttt{CalibPrompt}: Calibration-Aware Prompt Learning}
\label{subsection:calibprompt}

\begin{figure}[t]
    \centering
    \includegraphics[width=0.9\linewidth, trim=0cm 0cm 0cm 0cm, clip]{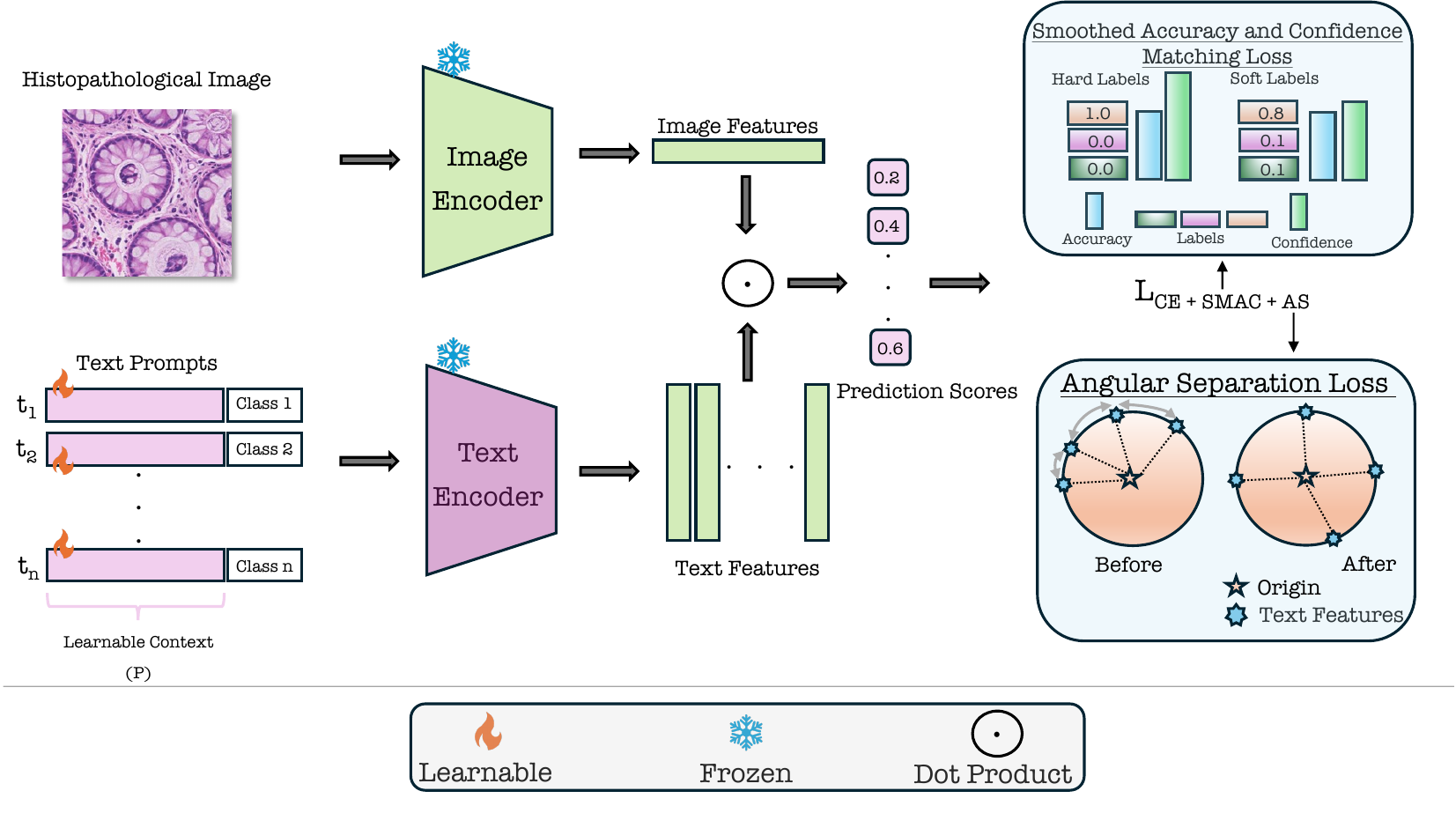}
    \caption{Overview of \texttt{CalibPrompt}. Learnable prompts are optimized using classification and calibration losses—\texttt{SMAC} and \texttt{AS}—while keeping the image and text encoders frozen. The \texttt{SMAC} loss aligns confidence with smoothed accuracy, while \texttt{AS} improves feature separation in the text embedding space.}
    \label{fig:main_fig_calibprompt}
\end{figure}

We introduce \texttt{CalibPrompt} as shown in Fig. \ref{fig:main_fig_calibprompt}, a new approach to improve confidence calibration in zero-shot classifiers based on Med-VLMs.
Motivated by our observations, CalibPrompt incorporates learnable prompts into the text encoder and optimizes them with our proposed calibration-aware auxiliary losses to enforce appropriate confidence calibration while keeping the model backbone frozen.
Specifically, given a zero-shot classifier $f$ based on a pre-trained Med-VLM $(\mathbf{E}_{\text{image}},\mathbf{E}_{\text{text}})$ and a few labeled samples $\{(\mathbf{I}_{n}, y_{n})\}_{n=1}^N$ from a downstream dataset $\mathcal{D}$, where $\mathbf{I}_n \in \mathcal{I}$ and $y_n \in \mathcal{Y}$, \texttt{CalibPrompt} optimizes the learnable prompts $\mathcal{P}$ to jointly minimize classification loss and calibration error:

\begin{align}
\label{eq:calibprompt}
    \mathcal{P}^{\ast} = \argmin_{\mathcal{P}} \frac{1}{N} \sum_{n=1}^{N} \Big[ 
    \mathcal{L}_{\text{CE}}(f_{\mathcal{P}}(\mathbf{I}_n), y_n) 
    + \lambda \mathcal{L}_{\text{calib}}(f_{\mathcal{P}}(\mathbf{I}_n), y_n) \Big],
\end{align}

\noindent where $\mathcal{L}_{\text{CE}}$ is the cross-entropy loss, $\mathcal{L}_{\text{calib}}$ is our overall calibration objective, and $\lambda$ balances accuracy and calibration objectives. The prompts are updated via backpropagation while keeping the Med-VLM frozen, preserving its pre-trained knowledge while optimizing for calibrated predictions.
To address miscalibration, we introduce two complementary objectives: conforming softened accuracy with predicted confidences ($\mathcal{L}_{\text{SMAC}}$) and the Angular Separation Loss ($\mathcal{L}_{\text{AS}}$), forming our calibration objective $\mathcal{L}_{\text{calib}} = \alpha \mathcal{L}_{\text{SMAC}} + \beta \mathcal{L}_{\text{AS}}$.

\noindent\textbf{\underline{Smoothed Accuracy and Confidence Matching (\texttt{SMAC})}:} Medical image classification often involves inherent class ambiguities, where diagnostic categories exhibit overlapping visual features. Traditional hard-label-based calibration methods~\cite{kumar2018trainable} enforce overly rigid decision boundaries, leading to miscalibrated overconfidence. To address this, we propose aligning predicted confidences with smoothed empirical class frequencies in a class-wise manner, termed \texttt{SMAC}. This provides a nuanced training signal that captures inherent ambiguities in medical imaging. Let $\mathbf{p}_n = f_{\mathcal{P}}(\mathbf{I}_n)$ denote the predicted probability distribution for image $\mathbf{I}_n$ across $K$ classes, and let $y_n \in \{1,2,...,K\}$ be the ground truth label. The \texttt{SMAC} loss is formulated as:
\begin{equation}
\mathcal{L}_{\text{SMAC}} = \frac{1}{K}\sum_{c=1}^{K}\big|\underbrace{\frac{1}{N}\sum_{n=1}^{N}p_n^{(c)}}_{\substack{\text{avg. predicted}\\\text{confidence}}}-\underbrace{\left[(1 - \alpha)f_c + \frac{\alpha(1 - f_c)}{K - 1}\right]}_{\substack{\text{smoothed class}\\\text{frequency}}}\big|,
\end{equation}
where $f_c=\frac{1}{N}\sum_{n=1}^{N}\mathbb{I}[y_n=c]$, and $\alpha\in[0,1)$ controls smoothing intensity. Using smoothed frequencies for confidence estimation, \texttt{SMAC} allows relaxed matching between predicted and empirical class distributions, thus reducing the likelihood of overconfident predictions in ambiguous scenarios.

\noindent\textbf{\underline{Angular Separation (\texttt{AS}) Loss }:}  
Building on our observation that prompt tuning increases text embedding similarity (see Fig.~\ref{fig:fracCE+SLMDCA+angle}), we address a key challenge in medical image classification where high inter-class feature similarity leads to overconfident predictions and degraded calibration. We propose an \textbf{Angular Separation Loss} for the textual embeddings, which discourages excessive similarity between class embeddings by minimizing their average pairwise cosine similarity. This ensures well-separated textual feature representations, improving confidence calibration while preserving classification accuracy. Mathematically, let 
$\textbf{Z} \in \mathbb{R}^{K \times D}$ be the text feature matrix, where each row $\textbf{z}_i$ represents the feature embedding of class $i$ in a D-dimensional space. We compute the cosine similarity matrix between all pairs of feature vectors as $\textbf{S} = \textbf{ZZ}^T$ where $S_{ij}$ measures the cosine similarity between class embeddings $\textbf{z}_i$ and $\textbf{z}_j$. To focus only on inter-class relationships, we mask the diagonal elements (self-similarities) as $\textbf{S}_{\text{off-diag}} = \textbf{S} - \text{diag}(\textbf{S})$. The Angular Separation Loss is then defined as the mean similarity across all class pairs:
\begin{equation}
   \mathcal{L}_{\text{AS}} = \frac{1}{K(K-1)} \sum_{i=1}^{K} \sum_{j \neq i} S_{ij}.
\end{equation}
By minimizing this loss, class embeddings become more distinct, reducing confidence overestimation and improving calibration. While \texttt{SMAC} loss refines probability-space confidence calibration, $\mathcal{L}_{\text{AS}}$ explicitly regularizes the text embedding space, ensuring both feature separability and confidence reliability.

\section{Experiments}
\input{tables/zeroshot_medvlms}

\input{tables/histopath_main_table_1}
\input{tables/xray_main_table_2}
\input{tables/multiple_calibration_metrics}


\textbf{Datasets, baselines and implementation details:} We hypothesize that the full fine-tuning approach can lead to overfitting in large networks when training data is limited, resulting in suboptimal feature representations.
We evaluate our method with two regularizers on four Med-VLMs: PLIP \cite{PLIP}, QuiltNet \cite{QUILTNET}, MedCLIP \cite{MEDCLIP}, and BioMedCLIP \cite{BIOMEDCLIP}, using five downstream datasets: COVIDX \cite{COVID}, RSNA18 \cite{RSNA18}, KatherColon \cite{KATHER}, PanNuke \cite{PANNUKE}, and DigestPath \cite{da2022digestpath}.
All experiments are conducted on an NVIDIA RTX A6000 GPU with 48GB memory. Our baselines include cross-entropy (CE), focal loss (FL) (given in supplementary), and label smoothing (LS), along with their combinations with established calibration regularization techniques such as DCA \cite{liang2020improved}, MMCE \cite{pmlr-v80-kumar18a} , MDCA \cite{hebbalaguppe2022stitch}, ZS-Norm\cite{murugesan2024robust}, Penalty\cite{murugesan2024robust}, MbLS \cite{liu2022devil}, and LogitNorm \cite{wei2022mitigating}. We evaluated these against our proposed methods \texttt{SMAC} and \texttt{SMAC} with \texttt{AS} across both full model fine-tuning and prompt learning approaches. We used an 8-shot setting (8 images per class) for both variations, we used a learning rate of  $2 \times 10^{-7}$ for full model fine-tuning, while for prompt learning we used a learning rate of 0.002.  Detailed hyperparameters are discussed in the supplementary.

Performance of the Med-VLMs is measured using accuracy (ACC). Similarly, for calibration, Expectation Calibration Error~\cite{ECE} (ECE), Adaptive Calibration Error \cite{ACE} (ACE), Maximum Calibration Error \cite{ECE} (MCE), and Expectation Calibration Error - Kernel Density Estimates\cite{ECE-KDE} $ECE^{KDE}$.


\subsection{Results} As shown in Table~\ref{tab:zeroshot_results}, Med-VLMs demonstrate encouraging zero-shot capability on downstream medical tasks, confirming their potential utility in real-world clinical pipelines. However, these models consistently suffer from severe calibration errors, often assigning high confidence to incorrect predictions. This miscalibration is particularly concerning in the medical domain, where overconfident false predictions can compromise trust and safety. 
Table~\ref{tab:results_plip_quiltnet} provides a detailed comparison, showing that our method yield substantially better calibration while maintaining competitive or improved classification accuracy. In particular, LS-based methods achieve an average ECE of only 1.85\%, an improvement over the 8.49\% baseline. Importantly, the accuracy remains stable or slightly better across most evaluation settings, confirming that calibration-aware objectives do not trade off predictive performance.  
Similarly, Table~\ref{tab:biomedclip_medclip_results} demonstrates that our regularizer attains state-of-the-art (SOTA) or second-best calibration across both BioMedCLIP and MedCLIP on two independent datasets. These results highlight the generalizability of our approach across different medical VLM architectures and data distributions.  
Finally, Table~\ref{tab:calibration_results} reports results across multiple calibration metrics (ECE, MCE, Brier score), showing consistent improvements for both PLIP and BioMedCLIP. The improvements are not restricted to one metric but hold across the board, which underlines the robustness of our method. 

\subsection{Ablations}
\begin{figure}[!htp]
    \centering
    \includegraphics[width=\linewidth, trim=0cm 5cm 0cm 3cm, clip]{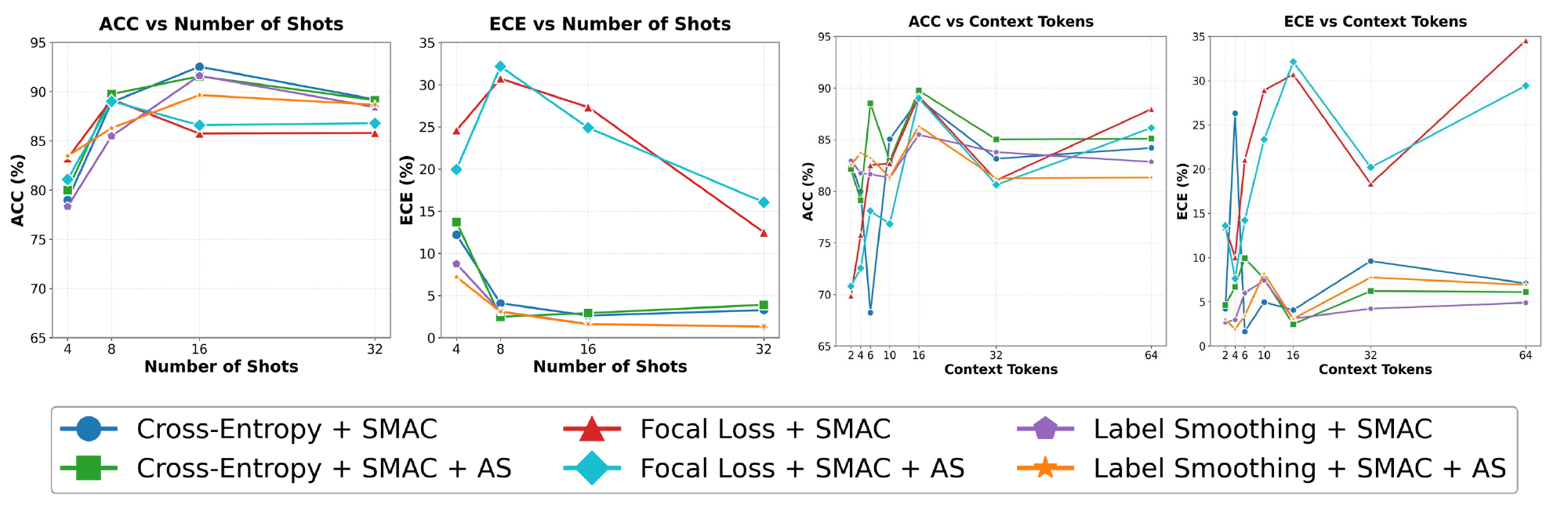}
    \caption{Comparative analysis of accuracy (ACC) and calibration error (ECE) for different loss functions
across varying few-shot counts and context token lengths.}
    \label{fig:num_shots}
\end{figure}

\textbf{Number of Few-shots:} Figure~\ref{fig:num_shots} illustrates the effect of varying the number of shots per class. As expected, increasing the number of training examples consistently improves accuracy (ACC) while also reducing calibration error (ECE). This suggests that additional supervision not only strengthens discriminative ability but also stabilizes confidence estimation.  

\noindent \textbf{Context Length:} Figure~\ref{fig:num_shots} also evaluates the influence of prompt token length. Our approach achieves optimal performance at 16 tokens, balancing expressivity and stability. Beyond this point, increasing the token count introduces instability and variance in both ACC and ECE.  

\noindent \textbf{Med-VLMs for Application-Specific Tasks:}  
We further examine the integration of our calibration-aware design with application-specific objectives. In particular, adding Angular Loss to PromptSmooth~\cite{hussein2024promptsmooth} yields measurable improvements in calibration while maintaining high accuracy. Specifically, PromptSmooth (few-shot + zero-shot) achieves 76.6\% ACC with 15.54\% ECE, while the addition of Angular Loss slightly decreases ACC to 76.2\% but significantly reduces ECE to 13.62\%. This reduction in calibration error, despite marginal accuracy changes, demonstrates that reliability can be substantially improved without compromising predictive utility.   

\section{Conclusion} 
We introduced \texttt{CalibPrompt}, a calibration-aware prompt tuning framework for Med-VLMs that improves confidence reliability while keeping the backbone frozen, making it efficient and deployment-friendly. By combining \texttt{SMCA} loss for probability-space calibration with our proposed \texttt{Angular Separation} (\texttt{AS}) loss for feature regularization, \texttt{CalibPrompt} mitigates overconfidence and enhances uncertainty estimation, yielding consistently lower calibration errors without sacrificing accuracy across multiple Med-VLMs and datasets. Extensive experiments and ablations confirm the complementary benefits of few-shot supervision and prompt design choices, while broader results highlight that calibration should be treated as an integral part of training rather than a post-hoc fix. In the future, we envision extending \texttt{CalibPrompt} beyond classification to more challenging tasks such as medical report generation, cross-modal retrieval, and multimodal reasoning, thereby advancing the development of calibration-aware Med-VLMs that are both reliable and clinically actionable.


\newpage
\bibliography{egbib}

\newpage

\section*{Supplementary Material - Calibration-Aware Prompt Learning for Medical Vision-Language Models}   
\addcontentsline{toc}{section}{Supplementary Material} 

\input{bmvc_appendix}

\end{document}

%% file: tables/zeroshot_medvlms.tex
\begin{table}[t]
\centering
\caption{Zero-shot accuracy (\%), confidence (\%), and ECE (\%) of Med-VLMs on X-ray and histopathology datasets. Final column shows over/underconfidence. Prompts used are shown above rows.}
\label{tab:zeroshot_results}
\resizebox{0.9\linewidth}{!}{
\begin{tabular}{llcccc}
\toprule
\rowcolor{gray!20}
\textbf{Dataset} & \textbf{Model} & \textbf{Accuracy (\%) $\uparrow$} & \textbf{Confidence (\%)} & \textbf{ECE (\%) $\downarrow$} & \textbf{Calibration} \\
\midrule
\rowcolor{gray!10} \multicolumn{6}{c}{\textbf{Hard Prompt:} \color{blue}\textit{A chest X-ray image of [class] patient}} \\
\midrule
\multirow{2}{*}{\textbf{COVID}} & BioMedCLIP & 84.37 & 95.0 & 10.70 & Overconfident \\
& MedCLIP & 78.77 & 50.0 & 28.67 & Underconfident \\
\midrule
\multirow{2}{*}{\textbf{RSNA18}} & BioMedCLIP & 49.71 & 79.0 & 29.49 & Overconfident \\
& MedCLIP & 47.60 & 34.0 & 14.05 & Underconfident \\
\midrule
\rowcolor{gray!10} \multicolumn{6}{c}{\textbf{Hard Prompt:} \color{purple}\textit{An H\&E image of [class]}} \\
\midrule
\multirow{2}{*}{\textbf{Kather}} & PLIP & 57.80 & 74.0 & 16.32 & Overconfident \\
& QuiltNet & 60.39 & 58.0 & 4.20 & Underconfident \\
\midrule
\rowcolor{gray!10} \multicolumn{6}{c}{\textbf{Hard Prompt:} \color{OrangeRed}\textit{An H\&E image patch of [class] skin tissue}} \\
\midrule
\multirow{2}{*}{\textbf{PanNuke}} & PLIP & 56.42 & 76.0 & 19.33 & Overconfident \\
& QuiltNet & 55.59 & 79.0 & 23.89 & Overconfident \\
\midrule
\rowcolor{gray!10} \multicolumn{6}{c}{\textbf{Hard Prompt:} \color{YellowOrange}\textit{An H\&E image patch of [class] tissue}} \\
\midrule
\multirow{2}{*}{\textbf{DigestPath}} & PLIP & 80.53 & 74.0 & 6.14 & Underconfident \\
& QuiltNet & 53.39 & 73.0 & 19.90 & Overconfident \\
\bottomrule
\end{tabular}
}
\end{table}

%% file: tables/histopath_main_table_1.tex
\begin{table}[t]
\begin{minipage}{\linewidth}
\centering
\caption{Comparison of proposed regularizers (\texttt{SMAC}, \texttt{SMAC+AS}) with baseline methods using Cross Entropy (CE), and Label Smoothing (LS). Accuracy (ACC, \%) and Expected Calibration Error (ECE, \%) are shown for PLIP and QuiltNet on histopathology datasets (Kather, PanNuke, DigestPath). Subscripts \textcolor{blue}{FT} and \textcolor{purple}{PL} denote Few-shot Fine-Tuning and Prompt Learning. Best results are in \textbf{bold}, second-best \underline{underlined}.}
\label{tab:results_plip_quiltnet}
\setlength{\tabcolsep}{1.8pt}
\resizebox{\linewidth}{!}{%
\begin{tabular}{l cccccc  cccccc cc}
\toprule
\rowcolor{gray!20} Model $\rightarrow$ & \multicolumn{6}{c}{\textbf{PLIP}} & \multicolumn{6}{c}{\textbf{QuiltNet}} & \multicolumn{2}{c}{\textbf{Average}}\\
\cmidrule(lr{3pt}){2-7} \cmidrule(lr{3pt}){8-13} \cmidrule(lr{3pt}){14-15}
\rowcolor{gray!20} Dataset $\rightarrow$ & \multicolumn{2}{c}{Kather} & \multicolumn{2}{c}{PanNuke} & \multicolumn{2}{c}{DigestPath}   & \multicolumn{2}{c}{Kather}  & \multicolumn{2}{c}{PanNuke} & \multicolumn{2}{c}{DigestPath} & \multicolumn{2}{c}{All} \\
\rowcolor{gray!20}Loss $\downarrow$ & ACC $\uparrow$ & ECE $\downarrow$ & ACC $\uparrow$ & ECE $\downarrow$ & ACC $\uparrow$ & ECE $\downarrow$ & Acc $\uparrow$ & ECE $\downarrow$ & ACC $\uparrow$ & ECE $\downarrow$ & ACC $\uparrow$ & ECE $\downarrow$ & ACC $\uparrow$ & ECE $\downarrow$\\ 
\midrule
\rowcolor{gray!20} \multicolumn{15}{c}{\textbf{Cross Entropy-based Losses}} \\
\cmidrule{1-15}
\rowcolor{white!10} Cross Entropy Loss$_{\textcolor{blue}{FT}}$ & 81.17 & 8.46  & 60.08 & 15.79 & 82.68 & 5.50  & 79.19 & 17.20  & 72.80 & 7.67 & 75.93 & 4.62 & 75.31 & 9.87  \\
\rowcolor{white!10} Cross Entropy Loss$_{\textcolor{purple}{PL}}$ & 83.91 & 5.92  & 66.70 & 17.82 & 82.87 & 9.50  & 87.97 & 2.49  & 69.82 & 19.70 & 81.59 & 11.27 & 78.81 & 11.12 \\
\rowcolor{white!10} CE + DCA$_{\textcolor{blue}{FT}}$  & 82.63 & 8.20  & 59.37 & 17.57 & 83.76 & 5.53 & 80.11 & 17.90 & 51.19 & 13.07 & 31.49 & 33.14 & 64.76 & 15.90 \\
\rowcolor{white!10} CE + DCA$_{\textcolor{purple}{PL}}$ & 85.74 & 3.60  & 67.94 & 11.48 & 74.71 & 13.91 & 88.16 & 1.50 & 74.47 & 4.29 & 84.24 & 1.71 & 79.21 & 6.08 \\
\rowcolor{white!10} CE + MMCE$_{\textcolor{blue}{FT}}$ & 81.16 & 8.46  & 60.15 & 15.68  & 82.86 & 5.52  & 79.07 & 17.06  & 73.05 & 7.26 & 80.50 & 7.60 & 76.13 & 10.26  \\
\rowcolor{white!10} CE + MMCE$_{\textcolor{purple}{PL}}$ & 83.52 & 3.07  & 67.05 & 12.16  & 78.81 & 9.68  & 90.97 & 2.11  & 68.08 & 17.95 & 85.13 & 4.49 & 78.93 & 8.24 \\
\rowcolor{white!10} CE + MDCA$_{\textcolor{blue}{FT}}$ & 81.14 & 8.41  & 59.90 & 16.08  & 82.54 & 5.48  & 79.14 & 17.17  & 72.80 & 7.67 & 31.07 & 32.53 & 67.77 & 14.56  \\
\rowcolor{white!10} CE + MDCA$_{\textcolor{purple}{PL}}$ & 83.91 & 5.79  & 70.67 & 9.69  & 88.92 & 4.08  & 89.99 & 1.61  & 69.29 & 13.46 & 84.89 & 6.55 & 81.28 & 6.86 \\
\rowcolor{white!10} MbLS$_{\textcolor{purple}{PL}}$ & 84.39 & 3.57  & 66.70 & 17.82  & 82.76 & 9.53  & 84.76 & 3.48  & 65.54 & 23.05 & 82.58 & 10.90 & 77.79 & 11.39  \\
\rowcolor{white!10} LogitNorm$_{\textcolor{purple}{PL}}$ & 86.52 & 5.12  & 57.10 & 31.72  & 84.80 & 9.04  & 88.22 & 3.42  & 72.91 & 13.88 & 87.17 & 5.26 & 79.45 & 11.41  \\
\rowcolor{white!10} ZS-Norm$_{\textcolor{purple}{PL}}$ & 85.63 & 3.07  & 71.52 & 17.22  & 82.84 & 7.31  & 91.91 & 0.87  & 69.55 & 19.18 & 84.78 & 6.37 & 81.04 & 9.00  \\
\rowcolor{white!10} Penalty$_{\textcolor{purple}{PL}}$ & 86.48 & 3.90  & 70.49 & 3.11  & 69.61 & \textbf{2.40}  & 89.29 & 12.28  & 59.88 & 3.41 & 79.23 & 17.53 & 75.83 & 7.11  \\
\midrule
\rowcolor{green!10} \textbf{CE + \texttt{SMAC}}$_{\textcolor{blue}{FT}}$ & 81.39 & 8.52 & 64.37 & 9.20 & 82.94 & 5.75 & 79.75 & 17.85 & 72.78 & 5.87 & 80.96 & 8.64 & 77.03 & 9.31 \\
\rowcolor{green!10} \textbf{CE + \texttt{SMAC}}$_{\textcolor{purple}{PL}}$ & 84.11 & 5.42 & 70.67 & 9.69 & 88.92 & 4.08 & 89.99 & 1.57 & 69.29 & 13.46 & 84.89 & 6.55 & 81.31 & 6.80 \\
\rowcolor{green!10} \textbf{CE + (\texttt{SMAC} + \texttt{AS})}$_{\textcolor{blue}{FT}}$ & 81.45 & 8.55 & 64.16 & 9.56 & 83.04 & 5.78 & 78.38 & 13.13 & 72.91 & 6.02 & 81.26 & 8.81 & 76.87 & 8.64 \\
\rowcolor{green!10} \textbf{CE + (\texttt{SMAC} + \texttt{AS})}$_{\textcolor{purple}{PL}}$ & 84.65 & 5.05 & 65.81 & 9.30 & 89.75 & \textbf{2.47} & 89.53 & 2.82 & 69.02 & 16.51 & 86.52 & 6.03 & 80.88 & 7.03 \\
\midrule
\rowcolor{gray!20} \multicolumn{15}{c}{\textbf{Label Smoothing-based Losses}} \\
\cmidrule{1-15}
\rowcolor{white!10} Label Smoothing$_{\textcolor{blue}{FT}}$ & 80.64 & 9.32  & 59.83 & 15.66 & 82.24 & 5.81  & 77.99 & 17.34  & 72.73 & 7.73 & 76.58 & 7.11 & 75.00 & 10.50  \\
\rowcolor{white!10} Label Smoothing$_{\textcolor{purple}{PL}}$ & 85.28 & 2.23  & 67.85 & 16.01 & 80.87 & 5.50  & 89.43 & 3.26  & 67.55 & 16.93 & 76.69 & 7.03 & 77.95 & 8.49 \\
\rowcolor{white!10} LS + MDCA$_{\textcolor{blue}{FT}}$ & 80.72 & 9.44  & 60.24 & 14.92  & 82.55 & 5.88  & 78.04 & 17.36  & 72.73 & 7.73 & 77.45 & 12.38 & 75.29 & 11.29  \\
\rowcolor{white!10} LS + MDCA$_{\textcolor{purple}{PL}}$ & 83.58 & \textbf{1.06}  & 68.93 & 9.39  & 83.29 & 3.49  & 91.53 & 4.37  & 75.32 & 3.91  & 88.05 & \underline{0.82} & 81.78 & 3.84 \\
\midrule
\rowcolor{green!10} \textbf{LS + \texttt{SMAC}}$_{\textcolor{blue}{FT}}$ & 80.91 & 9.47 & 65.10 & 6.13 & 83.15 & 6.10 & 79.48 & 17.83 & 72.20 & 5.29 & 76.88 & 6.25 & 76.29 & 8.51 \\
\rowcolor{green!10} \textbf{LS + \texttt{SMAC}}$_{\textcolor{purple}{PL}}$ & 83.59 & \underline{1.30} & 59.49 & \underline{2.38} & 85.48 & 3.11 & 90.57 & \textbf{0.89} & 69.64 & \underline{2.58} & 88.05 & \underline{0.82} & 79.47 & \textbf{1.85} \\
\rowcolor{green!10} \textbf{LS + (\texttt{SMAC} + \texttt{AS})}$_{\textcolor{blue}{FT}}$ & 80.95 & 9.48 & 64.67 & 6.78 & 83.06 & 6.04 & 77.98 & 12.99 & 72.09 & 5.15 & 76.68 & 6.12 & 75.91 & 7.76  \\
\rowcolor{green!10} \textbf{LS + (\texttt{SMAC} + \texttt{AS})}$_{\textcolor{purple}{PL}}$ & 85.08 & 3.11 & 58.68 & \textbf{2.19} & 86.30 & 3.08 & 87.52 & 3.58 & 71.93 & \underline{2.47} & 85.52 & \textbf{0.77} & 79.17 & \underline{2.53} \\
\bottomrule
\end{tabular}
}
\end{minipage}
\end{table}

%% file: tables/xray_main_table_2.tex
\begin{table}[t]
\begin{minipage}{\linewidth}
\centering
\caption{Comparison of our proposed calibration regularizers (\texttt{SMAC}, \texttt{SMAC+AS}) with baseline methods using Cross Entropy (CE), and Label Smoothing (LS). Results show Accuracy (ACC, \%) and Expected Calibration Error (ECE, \%) for BioMedCLIP and MedCLIP on COVID and RSNA datasets. Best results are in \textbf{bold}, second-best are \underline{underlined}.}
\label{tab:biomedclip_medclip_results}
\setlength{\tabcolsep}{1.8pt}
\resizebox{\linewidth}{!}{%
\begin{tabular}{l cccc cccc cc}
\toprule
\rowcolor{gray!20} Model $\rightarrow$ & \multicolumn{4}{c}{\textbf{BioMedCLIP}} & \multicolumn{4}{c}{\textbf{MedCLIP}} & \multicolumn{2}{c}{\textbf{Average}}\\
\cmidrule(lr{3pt}){2-5} \cmidrule(lr{3pt}){6-9} \cmidrule(lr{3pt}){10-11}
\rowcolor{gray!20} Dataset $\rightarrow$ & \multicolumn{2}{c}{COVID} & \multicolumn{2}{c}{RSNA} & \multicolumn{2}{c}{COVID} & \multicolumn{2}{c}{RSNA} & \multicolumn{2}{c}{All} \\
\rowcolor{gray!20}Loss $\downarrow$ & ACC $\uparrow$ & ECE $\downarrow$ & ACC $\uparrow$ & ECE $\downarrow$ & ACC $\uparrow$ & ECE $\downarrow$ & ACC $\uparrow$ & ECE $\downarrow$ & ACC $\uparrow$ & ECE $\downarrow$ \\
\midrule
\rowcolor{gray!20} \multicolumn{11}{c}{\textbf{Cross Entropy-based Losses}} \\
\rowcolor{white!10} Cross Entropy Loss$_{\textcolor{purple}{PL}}$ & 80.10 & 6.61 & 62.98 & 7.02 & 77.61 & 27.51 & 50.68 & 17.21 & 67.84 & 14.59 \\
\rowcolor{white!10} CE + DCA$_{\textcolor{purple}{PL}}$ & 59.79 & \underline{3.75} & 51.71 & 5.79 & 78.26 & 28.15 & 50.66 & 17.20 & 60.11 & 13.72 \\
\rowcolor{white!10} CE + MMCE$_{\textcolor{purple}{PL}}$ & 81.24 & 5.38 & 63.40 & 11.72 & 77.59 & 27.49 & 50.64 & 17.17 & 68.22 & 15.44 \\
\rowcolor{white!10} CE + MDCA$_{\textcolor{purple}{PL}}$ & 78.39 & \textbf{1.13} & 62.12 & 8.63 & 77.61 & 27.51 & 50.69 & 17.22 & 67.20 & 13.62 \\
\rowcolor{white!10} MbLS$_{\textcolor{purple}{PL}}$ & 80.10 & 6.61 & 62.22 & 6.97 & 77.61 & 27.51 & 50.68 & 17.21 & 67.65 & 14.58 \\
\rowcolor{white!10} LogitNorm$_{\textcolor{purple}{PL}}$ & 72.45 & 14.39 & 46.23 & 30.37 & 77.82 & 27.72 & 50.90 & 17.42 & 61.85 & 22.48 \\
\rowcolor{white!10} ZS-Norm$_{\textcolor{purple}{PL}}$ & 79.85 & 4.16 & 62.93 & 8.09 & 77.62 & 27.52 & 50.80 & 17.32 & 67.80 & 14.27 \\
\rowcolor{white!10} Penalty$_{\textcolor{purple}{PL}}$ & 77.85 & 9.35 & 50.54 & \textbf{2.40} & 77.55 & \underline{27.45} & 50.69 & 17.22 & 64.16 & 14.11 \\
\rowcolor{green!10} \textbf{CE + \texttt{SMAC}}$_{\textcolor{purple}{PL}}$ & 78.39 & \textbf{1.13} & 61.90 & 8.44 & 77.61 & 27.51 & 50.69 & 17.22 & 67.15 & \underline{13.58} \\
\rowcolor{green!10} \textbf{CE + (\texttt{SMAC} + \texttt{AS})}$_{\textcolor{purple}{PL}}$ & 74.88 & 4.04 & 58.75 & \underline{4.23} & 77.58 & 27.48 & 50.68 & 17.21 & 65.47 & \textbf{13.24} \\
\midrule
\rowcolor{gray!20} \multicolumn{11}{c}{\textbf{Label Smoothing-based Losses}} \\
\rowcolor{white!10} Label Smoothing$_{\textcolor{purple}{PL}}$ & 78.64 & 12.65 & 62.13 & 16.93 & 77.61 & 27.51 & 50.63 & \underline{17.16} & 67.25 & 18.56 \\
\rowcolor{white!10} LS + MDCA$_{\textcolor{purple}{PL}}$ & 74.77 & 13.33 & 63.13 & 18.89 & 77.51 & \textbf{27.41} & 50.63 & \underline{17.16} & 66.51 & 19.20 \\
\rowcolor{green!10} \textbf{LS + \texttt{SMAC}}$_{\textcolor{purple}{PL}}$ & 78.29 & 9.83 & 63.67 & 15.82 & 77.51 & \textbf{27.41} & 50.63 & \underline{17.16} & 67.53 & 17.56 \\
\rowcolor{green!10} \textbf{LS + (\texttt{SMAC} + \texttt{AS})}$_{\textcolor{purple}{PL}}$ & 74.84 & 4.69 & 59.51 & 5.74 & 77.51 & \textbf{27.41} & 50.61 & \textbf{17.14} & 65.62 & 13.75 \\
\bottomrule
\end{tabular}
}
\end{minipage}
\end{table}

%% file: tables/multiple_calibration_metrics.tex
\begin{table}[t]
\begin{minipage}{\linewidth}
\centering
\caption{Average calibration results of PLIP (PanNuke, DigestPath) and BioMedCLIP (COVID, RSNA) using ECE, ACE, MCE, and $\text{ECE}^{\text{KDE}}$. \textbf{Bold} and \underline{underline} denote best and second-best scores, respectively.}
\label{tab:calibration_results}
\setlength{\tabcolsep}{2pt}
\resizebox{0.9\linewidth}{!}{%
\begin{tabular}{l cccc|cccc}
\toprule
\rowcolor{gray!20} Model $\rightarrow$ & \multicolumn{4}{c|}{\textbf{PLIP}} & \multicolumn{4}{c}{\textbf{BioMedCLIP}} \\
\cmidrule(lr{3pt}){2-5} \cmidrule(lr{3pt}){6-9}
\rowcolor{gray!20}Loss $\downarrow$ & ECE $\downarrow$ & ACE $\downarrow$ & MCE $\downarrow$ & KDE $\downarrow$ & ECE $\downarrow$ & ACE $\downarrow$ & MCE $\downarrow$ & KDE $\downarrow$ \\ 
\midrule
\rowcolor{gray!20} \multicolumn{9}{c}{\textbf{Cross Entropy-based Losses}} \\
\rowcolor{white!10} Cross Entropy Loss$_{\textcolor{purple}{PL}}$ & 13.66 & 13.66 & 8.53 & 13.25 & 6.82 & 6.78 & 2.65 & 6.54 \\
\rowcolor{white!10} CE + DCA$_{\textcolor{purple}{PL}}$ & 12.70 & 12.90 & 7.40 & 12.76 & 4.77 & 7.03 & 2.26 & 4.43 \\
\rowcolor{white!10} CE + MMCE$_{\textcolor{purple}{PL}}$ & 10.92 & 11.06 & 6.33 & 10.77 & 8.55 & 8.55 & 3.57 & 8.49 \\
\rowcolor{white!10} CE + MDCA$_{\textcolor{purple}{PL}}$ & 6.89 & 6.88 & 4.09 & 6.60 & 4.88 & 4.87 & 2.08 & 5.10 \\
\rowcolor{white!10} MbLS$_{\textcolor{purple}{PL}}$ & 13.68 & 13.68 & 8.53 & 13.26 & 6.79 & 6.77 & 2.65 & 6.56 \\
\rowcolor{white!10} LogitNorm$_{\textcolor{purple}{PL}}$ & 20.38 & 20.38 & 14.46 & 19.73 & 22.38 & 22.38 & 10.47 & 22.18 \\
\rowcolor{white!10} ZS-Norm$_{\textcolor{purple}{PL}}$ & 12.27 & 12.27 & 7.99 & 11.82 & 6.13 & 6.11 & 2.11 & 5.91 \\
\rowcolor{white!10} Penalty$_{\textcolor{purple}{PL}}$ & 2.76 & \textbf{3.05} & \textbf{1.16} & 3.27 & 5.88 & 6.06 & 2.26 & 6.08 \\
\rowcolor{green!10} \textbf{CE + \texttt{SMAC}}$_{\textcolor{purple}{PL}}$ & 6.89 & 6.88 & 4.09 & 6.60 & \underline{4.79} & \underline{4.78} & \underline{2.09} & \underline{5.02} \\
\rowcolor{green!10} \textbf{CE + (\texttt{SMAC} + \texttt{AS})}$_{\textcolor{purple}{PL}}$ & 5.89 & 5.77 & 2.90 & 5.82 & \textbf{4.14} & \textbf{4.28} & \textbf{1.70} & \textbf{4.60} \\
\midrule
\rowcolor{gray!20} \multicolumn{9}{c}{\textbf{Label Smoothing-based Losses}} \\
\rowcolor{white!10} Label Smoothing$_{\textcolor{purple}{PL}}$ & 10.76 & 10.76 & 6.32 & 10.71 & 14.79 & 14.79 & 7.08 & 14.64 \\
\rowcolor{white!10} LS + MDCA$_{\textcolor{purple}{PL}}$ & 6.44 & 6.43 & 2.87 & 6.37 & 16.09 & 16.09 & 9.19 & 15.87 \\
\rowcolor{green!10} \textbf{LS + \texttt{SMAC}}$_{\textcolor{purple}{PL}}$ & \underline{2.75} & 4.05 & 1.90 & \textbf{2.09} & 12.82 & 12.82 & 5.49 & 12.69 \\
\rowcolor{green!10} \textbf{LS + (\texttt{SMAC} + \texttt{AS})}$_{\textcolor{purple}{PL}}$ & \textbf{2.64} & \underline{3.51} & \underline{1.73} & \underline{2.25} & 5.22 & 5.14 & 2.15 & 5.27 \\
\bottomrule
\end{tabular}
}
\end{minipage}
\end{table}

%% file: bmvc_appendix.tex

\maketitle

\section{Hyperparameter Settings}
Our hyperparameter selection revealed distinct patterns across imaging modalities and model architectures. Tables~\ref{tab:histo_params} and~\ref{tab:radio_params} summarize our key hyperparameter configurations.

For histopathology datasets, we observed that effective calibration required more conservative regularization, with smaller \texttt{SMAC} values ($\alpha \approx 0.03$-$0.07$) and minimal \texttt{AS} weights (0.001-0.01) for most models. This suggests that histopathology models are more sensitive to excessive regularization, likely due to the fine-grained nature of cellular features that require preserving subtle discriminative patterns. Interestingly, the QuiltNet architecture demonstrated greater robustness to stronger regularization, particularly on the Kather dataset where an \texttt{AS} weight of 1.0 proved optimal.

In contrast, radiological datasets consistently benefited from stronger regularization, with higher \texttt{SMAC} values ($\alpha = 0.1$-$0.2$) and substantially larger \texttt{AS} weights (1.0-3.0). This difference suggests that models trained on radiological images are more prone to overconfidence, potentially due to the lower spatial resolution of diagnostic features, necessitating stronger calibration signals during training.

We further extended our analysis to include advanced calibration techniques such as Margin-based Logit Suppression (MBLS) and LogitNorm. For MBLS, we found that a consistent margin threshold of 10.0 with an $\alpha$ value of 0.1 worked effectively across both imaging modalities. The consistency of these parameters, unlike the variability observed with \texttt{SMAC} and \texttt{AS}, suggests that MBLS provides a more universally applicable calibration mechanism that is less sensitive to dataset-specific characteristics. LogitNorm similarly demonstrated robust performance with a standard temperature parameter of 1.0 across all datasets, indicating its adaptability across different medical imaging contexts.

For zero-shot calibration techniques (ECCV\_PENALTY and ECCV\_ZS), we maintained uniform weighting factors of 1.0 across all datasets. While these methods offer the advantage of leveraging pre-trained knowledge, their consistent parameterization reflects their primary reliance on the underlying zero-shot model rather than dataset-specific hyperparameter tuning.

\input{tables/histo_hyperparam}
\input{tables/xray_hyperparam}

For combined loss functions, we maintained consistency in hyperparameter values. When using \texttt{SMAC} with Label Smoothing (LS), we ensured matching $\alpha$ values (e.g., LS $\alpha$=0.05 with SMAC $\alpha$=0.05 for Kather). Similarly, when combining with Focal Loss (FL), we maintained consistent hyperparameters across loss components (e.g., FL $\gamma$=3.0 with SMAC $\alpha$=0.1 for radiological datasets). This consistency principle was extended to our newer calibration techniques as well, particularly when combining MBLS with other losses, where we maintained its standard parameterization to preserve its inherent calibration capabilities.

For combined loss functions, we maintained consistency in hyperparameter values. When using \texttt{SMAC} with Label Smoothing (LS), we ensured matching $\alpha$ values (e.g., LS $\alpha$=0.05 with SMAC $\alpha$=0.05 for Kather). Similarly, when combining with Focal Loss (FL), we maintained consistent hyperparameters across loss components (e.g., FL $\gamma$=3.0 with SMAC $\alpha$=0.1 for radiological datasets).

\section{Evaluation Metrics}
We evaluated model performance using both accuracy and calibration metrics:
\begin{itemize}
    \item \textbf{Accuracy (ACC)}: Proportion of correctly classified samples, indicating overall predictive performance.

    \item \textbf{Expected Calibration Error (ECE)} \cite{ECE}: Measures the weighted average difference between predicted confidence and actual accuracy across 10 fixed-width bins.

    \item \textbf{Adaptive Calibration Error (ACE)} \cite{ACE}: Similar to ECE but uses adaptive binning to ensure equal sample counts per bin, improving reliability on skewed confidence distributions.

    \item \textbf{Maximum Calibration Error (MCE)} \cite{ECE}: Reports the largest absolute difference between confidence and accuracy across all bins, highlighting worst-case miscalibration.

    \item \textbf{ECE with Kernel Density Estimation ($\text{ECE}^{\text{KDE}}$)} \cite{ECE-KDE}: A binning-free metric that uses kernel density estimation to compute a continuous calibration error estimate. Our implementation uses a simplified Gaussian kernel rather than the Dirichlet/Beta kernels from the original paper, offering computational efficiency while maintaining effectiveness for medical imaging tasks. We employ adaptive bandwidth selection (ranging from 0.05 to 0.3) based on dataset size, which is particularly beneficial for few-shot learning scenarios where test sets can vary significantly in size. This approach avoids binning artifacts and provides a smoother, more continuous estimate of the confidence-accuracy relationship.
\end{itemize}

\section{Focal Loss Experiments}
\label{sec:focal_loss}

In this supplementary material, we present additional experiments using focal loss that were not included in the main paper. These experiments complement our main findings and provide further insights into calibration techniques for vision-language models.

\subsection{Experimental Setup}

We conducted experiments applying focal loss to our context optimization approach across various datasets. The focal loss function is defined as:

$$FL(p_t) = -(1-p_t)^\gamma \log(p_t)$$

where $p_t$ represents the model's predicted probability for the true class and $\gamma$ is the focusing parameter that determines how much to down-weight well-classified examples.

\subsection{Results}

The results of our focal loss experiments are presented in the following tables:

\input{tables/histopath_main_table_1_focal}
\input{tables/xray_main_table_2_focal}
\input{tables/multiple_calibration_metrics_focal}

Table~\ref{tab:results_plip_quiltnet_focal} shows the performance on histopathology datasets, while Table~\ref{tab:biomedclip_medclip_results_focal} presents results on X-ray datasets. 
Table~\ref{tab:calibration_results_focal} provides a comprehensive analysis using various calibration metrics beyond ECE.


%% file: tables/histo_hyperparam.tex
\begin{table}[ht]
\centering
\caption{Hyperparameter settings for histopathology datasets.}
\label{tab:histo_params}
\begin{tabular}{lccc}
\toprule
\textbf{Parameter} & \textbf{Kather} & \textbf{Pannuke} & \textbf{DigestPath} \\
\midrule
DCA weight & 9.0 & 9.0 & 9.0 \\
Focal Loss ($\gamma$) & 3.0 & 3.0 & 3.0 \\
Label Smoothing ($\alpha$) & 0.05 & 0.05 & 0.05 \\
MMCE weight & 1.0 & 1.0 & 1.0 \\
MDCA weight & 1.0 & 1.0 & 1.0 \\
\midrule
\multicolumn{4}{c}{\textbf{PLIP Model}} \\
\midrule
\texttt{SMAC} ($\alpha$) & 0.03-0.07 & 0.2 & 0.03 \\
\texttt{AS} weight & 0.01 & 0.1 & 0.001 \\
\midrule
\multicolumn{4}{c}{\textbf{QuiltNet Model}} \\
\midrule
\texttt{SMAC} ($\alpha$) & 0.01-0.07 & 0.1 & 0.05 \\
\texttt{AS} weight & 1.0 & 0.001 & 0.001 \\
\bottomrule
\end{tabular}
\end{table}

%% file: tables/xray_hyperparam.tex
\begin{table}[ht]
\centering
\caption{Hyperparameter settings for radiological datasets.}
\label{tab:radio_params}
\begin{tabular}{lcc}
\toprule
\textbf{Parameter} & \textbf{BioMedCLIP} & \textbf{MedCLIP} \\
\midrule
DCA weight & 9.0 & 9.0 \\
Focal Loss ($\gamma$) & 3.0 & 3.0 \\
Label Smoothing ($\alpha$) & 0.2 & 0.2 \\
MMCE weight & 2.0 & 2.0 \\
MDCA weight & 1.0 & 1.0 \\
\texttt{SMAC} ($\alpha$) & 0.1 & 0.2 \\
\texttt{AS} weight & 3.0 & 1.0 \\
\bottomrule
\end{tabular}
\end{table}

%% file: tables/histopath_main_table_1_focal.tex
\begin{table}[t]
\begin{minipage}{\linewidth}
\centering
\caption{Comparison of proposed regularizers (\texttt{SMAC}, \texttt{SMAC+AS}) with baseline methods using Focal Loss (FL). Accuracy (ACC, \%) and Expected Calibration Error (ECE, \%) are shown for PLIP and QuiltNet on histopathology datasets (Kather, PanNuke, DigestPath). Subscripts \textcolor{blue}{FT} and \textcolor{purple}{PL} denote Few-shot Fine-Tuning and Prompt Learning. Best results are in \textbf{bold}, second-best \underline{underlined}.}
\label{tab:results_plip_quiltnet_focal}
\setlength{\tabcolsep}{1.8pt}
\resizebox{0.85\linewidth}{!}{%
\begin{tabular}{l cccccc  cccccc cc}
\toprule
\rowcolor{gray!20} Model $\rightarrow$ & \multicolumn{6}{c}{\textbf{PLIP}} & \multicolumn{6}{c}{\textbf{QuiltNet}} & \multicolumn{2}{c}{\textbf{Average}}\\
\cmidrule(lr{3pt}){2-7} \cmidrule(lr{3pt}){8-13} \cmidrule(lr{3pt}){14-15}
\rowcolor{gray!20} Dataset $\rightarrow$ & \multicolumn{2}{c}{Kather} & \multicolumn{2}{c}{PanNuke} & \multicolumn{2}{c}{DigestPath}   & \multicolumn{2}{c}{Kather}  & \multicolumn{2}{c}{PanNuke} & \multicolumn{2}{c}{DigestPath} & \multicolumn{2}{c}{All} \\
\rowcolor{gray!20}Loss $\downarrow$ & ACC $\uparrow$ & ECE $\downarrow$ & ACC $\uparrow$ & ECE $\downarrow$ & ACC $\uparrow$ & ECE $\downarrow$ & Acc $\uparrow$ & ECE $\downarrow$ & ACC $\uparrow$ & ECE $\downarrow$ & ACC $\uparrow$ & ECE $\downarrow$ & ACC $\uparrow$ & ECE $\downarrow$\\ 
\midrule
\rowcolor{gray!20} \multicolumn{15}{c}{\textbf{Focal Loss-based Losses}} \\
\cmidrule{1-15}
\rowcolor{white!10} Focal Loss$_{\textcolor{blue}{FT}}$ & 80.91 & 12.06  & 58.23 & 17.00  & 80.56 & \textbf{6.22}  & 79.23 & 19.84  & 72.89 & 7.93 & 77.81 & 7.73 & 74.94 & 11.80  \\
\rowcolor{white!10} Focal Loss$_{\textcolor{purple}{PL}}$ & 84.07 & 2.82  & 62.65 & \textbf{3.27}  & 87.30 & 7.94  & 86.36 & \underline{2.02}  & 68.93 & 5.68  & 84.94 & \textbf{3.33} & 79.04 & \textbf{4.18} \\
\rowcolor{white!10} FL + MDCA$_{\textcolor{blue}{FT}}$ & 80.91 & 12.03  & 58.59 & 16.38  & 81.07 & \underline{6.50}  & 79.35 & 19.94  & 73.01 & 8.02 & 77.50 & 11.46 & 75.07 & 12.39 \\
\rowcolor{white!10} FL + MDCA$_{\textcolor{purple}{PL}}$ & 85.96 & 2.60  & 63.82 & 10.30  & 89.22 & 30.71  & 89.81 & 3.72  & 72.89 & 19.64  & 90.20 & 36.04 & 82.00 & 17.17 \\
\midrule
\rowcolor{green!10} \textbf{FL + \texttt{SMAC}}$_{\textcolor{blue}{FT}}$ & 81.32 & 12.18 & 62.72 & 9.74 & 82.82 & 7.58 & 79.78 & 20.37 & 72.85 & \underline{6.52} & 80.12 & 10.20 & 76.60 & 11.10 \\
\rowcolor{green!10} \textbf{FL + \texttt{SMAC}}$_{\textcolor{purple}{PL}}$ & 85.99 & \underline{2.59} & 63.82 & 10.30 & 89.22 & 30.71 & 87.05 & 2.43 & 72.89 & 19.64 & 90.20 & 36.04 & 81.53 & 16.95 \\
\rowcolor{green!10} \textbf{FL + (\texttt{SMAC} + \texttt{AS})}$_{\textcolor{blue}{FT}}$ & 81.32 & 12.18 & 62.47 & 10.14 & 82.65 & 7.50 & 78.61 & 16.23 & 72.64 & \textbf{6.32} & 74.24 & \underline{5.92} & 75.32 & \underline{9.72} \\
\rowcolor{green!10} \textbf{FL + (\texttt{SMAC} + \texttt{AS})}$_{\textcolor{purple}{PL}}$ & 85.21 & \textbf{2.58}  & 62.44 & \underline{8.95} & 89.02 & 32.15 & 88.97 & \textbf{1.27}  & 66.09 & 13.64 & 82.13 & 30.31 & 78.98 & 14.82 \\
\bottomrule
\end{tabular}
}
\end{minipage}
\end{table}

%% file: tables/xray_main_table_2_focal.tex
\begin{table}[t]
\begin{minipage}{\linewidth}
\centering
\caption{Comparison of our proposed calibration regularizers (\texttt{SMAC}, \texttt{SMAC+AS}) with baseline methods using Focal Loss (FL). Results show Accuracy (ACC, \%) and Expected Calibration Error (ECE, \%) for BioMedCLIP and MedCLIP on COVID and RSNA datasets. Best results are in \textbf{bold}, second-best are \underline{underlined}.}
\label{tab:biomedclip_medclip_results_focal}
\setlength{\tabcolsep}{1.8pt}
\resizebox{0.9\linewidth}{!}{%
\begin{tabular}{l cccc cccc cc}
\toprule
\rowcolor{gray!20} Model $\rightarrow$ & \multicolumn{4}{c}{\textbf{BioMedCLIP}} & \multicolumn{4}{c}{\textbf{MedCLIP}} & \multicolumn{2}{c}{\textbf{Average}}\\
\cmidrule(lr{3pt}){2-5} \cmidrule(lr{3pt}){6-9} \cmidrule(lr{3pt}){10-11}
\rowcolor{gray!20} Dataset $\rightarrow$ & \multicolumn{2}{c}{COVID} & \multicolumn{2}{c}{RSNA} & \multicolumn{2}{c}{COVID} & \multicolumn{2}{c}{RSNA} & \multicolumn{2}{c}{All} \\
\rowcolor{gray!20}Loss $\downarrow$ & ACC $\uparrow$ & ECE $\downarrow$ & ACC $\uparrow$ & ECE $\downarrow$ & ACC $\uparrow$ & ECE $\downarrow$ & ACC $\uparrow$ & ECE $\downarrow$ & ACC $\uparrow$ & ECE $\downarrow$ \\
\midrule
\rowcolor{gray!20} \multicolumn{11}{c}{\textbf{Focal Loss-based Losses}} \\
\rowcolor{white!10} Focal Loss$_{\textcolor{purple}{PL}}$ & 78.42 & 16.47 & 61.13 & \underline{17.45} & 77.55 & \underline{27.45} & 50.66 & \underline{17.19} & 66.94 & 19.64 \\
\rowcolor{white!10} FL + MDCA$_{\textcolor{purple}{PL}}$ & 71.18 & \underline{12.17} & 61.74 & 18.58 & 77.49 & \textbf{27.39} & 50.64 & \textbf{17.17} & 65.26 & \underline{18.83} \\
\rowcolor{green!10} \textbf{FL + \texttt{SMAC}}$_{\textcolor{purple}{PL}}$ & 71.18 & \underline{12.17} & 61.88 & 18.70 & 77.49 & \textbf{27.39} & 50.64 & \textbf{17.17} & 65.30 & 18.86 \\
\rowcolor{green!10} \textbf{FL + (\texttt{SMAC} + \texttt{AS})}$_{\textcolor{purple}{PL}}$ & 72.86 & \textbf{7.14} & 60.86 & \textbf{15.20} & 77.49 & \textbf{27.39} & 50.64 & \textbf{17.17} & 65.46 & \textbf{16.73} \\
\bottomrule
\end{tabular}
}
\end{minipage}
\end{table}

%% file: tables/multiple_calibration_metrics_focal.tex
\begin{table}[!htp]
\begin{minipage}{\linewidth}
\centering
\caption{Average calibration results of PLIP (PanNuke, DigestPath) and BioMedCLIP (COVID, RSNA) using ECE, ACE, MCE, and $\text{ECE}^{\text{KDE}}$. \textbf{Bold} and \underline{underline} denote best and second-best scores, respectively.}
\label{tab:calibration_results_focal}
\setlength{\tabcolsep}{2pt}
\resizebox{0.8\linewidth}{!}{%
\begin{tabular}{l cccc|cccc}
\toprule
\rowcolor{gray!20} Model $\rightarrow$ & \multicolumn{4}{c|}{\textbf{PLIP}} & \multicolumn{4}{c}{\textbf{BioMedCLIP}} \\
\cmidrule(lr{3pt}){2-5} \cmidrule(lr{3pt}){6-9}
\rowcolor{gray!20}Loss $\downarrow$ & ECE $\downarrow$ & ACE $\downarrow$ & MCE $\downarrow$ & KDE $\downarrow$ & ECE $\downarrow$ & ACE $\downarrow$ & MCE $\downarrow$ & KDE $\downarrow$ \\ 
\midrule
\rowcolor{gray!20} \multicolumn{9}{c}{\textbf{Focal Loss-based Losses}} \\
\rowcolor{white!10} Focal Loss$_{\textcolor{purple}{PL}}$ & \textbf{5.61} & \textbf{5.53} & \textbf{2.09} & \textbf{5.83} & 16.96 & 16.95 & 10.63 & 16.87 \\
\rowcolor{white!10} FL + MDCA$_{\textcolor{purple}{PL}}$ & \underline{20.51} & \underline{20.51} & \underline{12.98} & \underline{20.65} & \underline{15.38} & \underline{15.38} & \underline{9.52} & \underline{15.01} \\
\rowcolor{green!10} \textbf{FL + \texttt{SMAC}}$_{\textcolor{purple}{PL}}$ & \underline{20.51} & \underline{20.51} & \underline{12.98} & \underline{20.65} & 15.44 & 15.43 & 9.80 & 15.09 \\
\rowcolor{green!10} \textbf{FL + (\texttt{SMAC} + \texttt{AS})}$_{\textcolor{purple}{PL}}$ & 20.55 & 20.77 & 15.71 & 20.60 & \textbf{11.17} & \textbf{11.17} & \textbf{6.78} & \textbf{11.05} \\
\bottomrule
\end{tabular}
}
\end{minipage}
\end{table}

%% file: bmvc_final.bbl
\begin{thebibliography}{37}
\providecommand{\natexlab}[1]{#1}
\providecommand{\url}[1]{\texttt{#1}}
\expandafter\ifx\csname urlstyle\endcsname\relax
  \providecommand{\doi}[1]{doi: #1}\else
  \providecommand{\doi}{doi: \begingroup \urlstyle{rm}\Url}\fi

\bibitem[Chia et~al.(2024)Chia, Antaki, Zhou, Turner, Lee, and Keane]{chia2024foundation}
Mark~A Chia, Fares Antaki, Yukun Zhou, Angus~W Turner, Aaron~Y Lee, and Pearse~A Keane.
\newblock Foundation models in ophthalmology.
\newblock \emph{British Journal of Ophthalmology}, 108\penalty0 (10):\penalty0 1341--1348, 2024.

\bibitem[Da et~al.(2022)Da, Huang, Li, Zuo, Zhang, Liu, Chen, Li, Xu, Hu, et~al.]{da2022digestpath}
Qian Da, Xiaodi Huang, Zhongyu Li, Yanfei Zuo, Chenbin Zhang, Jingxin Liu, Wen Chen, Jiahui Li, Dou Xu, Zhiqiang Hu, et~al.
\newblock Digestpath: A benchmark dataset with challenge review for the pathological detection and segmentation of digestive-system.
\newblock \emph{Medical Image Analysis}, 80:\penalty0 102485, 2022.

\bibitem[Gamper et~al.(2019)Gamper, Alemi~Koohbanani, Benet, Khuram, and Rajpoot]{PANNUKE}
Jevgenij Gamper, Navid Alemi~Koohbanani, Ksenija Benet, Ali Khuram, and Nasir Rajpoot.
\newblock Pannuke: an open pan-cancer histology dataset for nuclei instance segmentation and classification.
\newblock In \emph{Digital Pathology: 15th European Congress, ECDP 2019, Warwick, UK, April 10--13, 2019, Proceedings 15}, pages 11--19. Springer, 2019.

\bibitem[Guo et~al.(2017)Guo, Pleiss, Sun, and Weinberger]{guo2017calibration}
Chuan Guo, Geoff Pleiss, Yu~Sun, and Kilian~Q Weinberger.
\newblock On calibration of modern neural networks.
\newblock In \emph{ICML}, pages 1321--1330. PMLR, 2017.

\bibitem[Hanif et~al.(2024)Hanif, Shamshad, Awais, Naseer, Khan, Nandakumar, Khan, and Anwer]{hanif2024baple}
Asif Hanif, Fahad Shamshad, Muhammad Awais, Muzammal Naseer, Fahad~Shahbaz Khan, Karthik Nandakumar, Salman Khan, and Rao~Muhammad Anwer.
\newblock Baple: Backdoor attacks on medical foundational models using prompt learning.
\newblock In \emph{International Conference on Medical Image Computing and Computer-Assisted Intervention}, pages 443--453. Springer, 2024.

\bibitem[Hebbalaguppe et~al.(2022{\natexlab{a}})Hebbalaguppe, Prakash, Madan, and Arora]{hebbalaguppe2022stitch}
Ramya Hebbalaguppe, Jatin Prakash, Neelabh Madan, and Chetan Arora.
\newblock A stitch in time saves nine: A train-time regularizing loss for improved neural network calibration.
\newblock In \emph{Proceedings of the IEEE/CVF Conference on Computer Vision and Pattern Recognition}, pages 16081--16090, 2022{\natexlab{a}}.

\bibitem[Hebbalaguppe et~al.(2022{\natexlab{b}})Hebbalaguppe, Prakash, Madan, and Arora]{mdca}
Ramya Hebbalaguppe, Jatin Prakash, Neelabh Madan, and Chetan Arora.
\newblock A stitch in time saves nine: A train-time regularizing loss for improved neural network calibration.
\newblock In \emph{Proceedings of the IEEE/CVF Conference on Computer Vision and Pattern Recognition}, pages 16081--16090, 2022{\natexlab{b}}.

\bibitem[Huang et~al.(2020)Huang, Li, Macheret, Gabriel, and Ohno-Machado]{huang2020tutorial}
Yingxiang Huang, Wentao Li, Fima Macheret, Rodney~A Gabriel, and Lucila Ohno-Machado.
\newblock A tutorial on calibration measurements and calibration models for clinical prediction models.
\newblock \emph{Journal of the American Medical Informatics Association}, 27\penalty0 (4):\penalty0 621--633, 2020.

\bibitem[Huang et~al.(2023)Huang, Bianchi, Yuksekgonul, Montine, and Zou]{PLIP}
Zhi Huang, Federico Bianchi, Mert Yuksekgonul, Thomas~J Montine, and James Zou.
\newblock A visual--language foundation model for pathology image analysis using medical twitter.
\newblock \emph{Nature medicine}, 29\penalty0 (9):\penalty0 2307--2316, 2023.

\bibitem[Hussein et~al.(2024)Hussein, Shamshad, Naseer, and Nandakumar]{hussein2024promptsmooth}
Noor Hussein, Fahad Shamshad, Muzammal Naseer, and Karthik Nandakumar.
\newblock Promptsmooth: Certifying robustness of medical vision-language models via prompt learning.
\newblock In \emph{International Conference on Medical Image Computing and Computer-Assisted Intervention}, pages 698--708. Springer, 2024.

\bibitem[Ikezogwo et~al.(2023)Ikezogwo, Seyfioglu, Ghezloo, Geva, Sheikh~Mohammed, Anand, Krishna, and Shapiro]{QUILTNET}
Wisdom Ikezogwo, Saygin Seyfioglu, Fatemeh Ghezloo, Dylan Geva, Fatwir Sheikh~Mohammed, Pavan~Kumar Anand, Ranjay Krishna, and Linda Shapiro.
\newblock Quilt-1m: One million image-text pairs for histopathology.
\newblock \emph{Advances in neural information processing systems}, 36:\penalty0 37995--38017, 2023.

\bibitem[Kather et~al.(2019)Kather, Krisam, Charoentong, Luedde, Herpel, Weis, Gaiser, Marx, Valous, Ferber, et~al.]{KATHER}
Jakob~Nikolas Kather, Johannes Krisam, Pornpimol Charoentong, Tom Luedde, Esther Herpel, Cleo-Aron Weis, Timo Gaiser, Alexander Marx, Nektarios~A Valous, Dyke Ferber, et~al.
\newblock Predicting survival from colorectal cancer histology slides using deep learning: A retrospective multicenter study.
\newblock \emph{PLoS medicine}, 16\penalty0 (1):\penalty0 e1002730, 2019.

\bibitem[Kugathasan et~al.(2024)Kugathasan, Zhou, Izzo, Kuruppu, Baliah, and Khan]{kugathasan2024matching}
Vinith Kugathasan, Honglu Zhou, Zachary Izzo, Gayal Kuruppu, Sanoojan Baliah, and Muhammad~Haris Khan.
\newblock Matching confidences and softened target occurrences for calibration.
\newblock In \emph{2024 International Conference on Digital Image Computing: Techniques and Applications (DICTA)}, pages 109--116. IEEE, 2024.

\bibitem[Kumar et~al.(2018{\natexlab{a}})Kumar, Sarawagi, and Jain]{kumar2018trainable}
Aviral Kumar, Sunita Sarawagi, and Ujjwal Jain.
\newblock Trainable calibration measures for neural networks from kernel mean embeddings.
\newblock In \emph{International Conference on Machine Learning}, pages 2805--2814. PMLR, 2018{\natexlab{a}}.

\bibitem[Kumar et~al.(2018{\natexlab{b}})Kumar, Sarawagi, and Jain]{pmlr-v80-kumar18a}
Aviral Kumar, Sunita Sarawagi, and Ujjwal Jain.
\newblock Trainable calibration measures for neural networks from kernel mean embeddings.
\newblock In Jennifer Dy and Andreas Krause, editors, \emph{Proceedings of the 35th International Conference on Machine Learning}, volume~80 of \emph{Proceedings of Machine Learning Research}, pages 2805--2814. PMLR, 10--15 Jul 2018{\natexlab{b}}.
\newblock URL \url{https://proceedings.mlr.press/v80/kumar18a.html}.

\bibitem[Lambert et~al.(2024)Lambert, Forbes, Doyle, Dehaene, and Dojat]{lambert2024trustworthy}
Benjamin Lambert, Florence Forbes, Senan Doyle, Harmonie Dehaene, and Michel Dojat.
\newblock Trustworthy clinical ai solutions: a unified review of uncertainty quantification in deep learning models for medical image analysis.
\newblock \emph{Artificial Intelligence in Medicine}, page 102830, 2024.

\bibitem[Liang et~al.(2020)Liang, Zhang, Wang, and Jacobs]{liang2020improved}
Gongbo Liang, Yu~Zhang, Xiaoqin Wang, and Nathan Jacobs.
\newblock Improved trainable calibration method for neural networks on medical imaging classification.
\newblock \emph{arXiv preprint arXiv:2009.04057}, 2020.

\bibitem[Liu et~al.(2022)Liu, Ben~Ayed, Galdran, and Dolz]{liu2022devil}
Bingyuan Liu, Ismail Ben~Ayed, Adrian Galdran, and Jose Dolz.
\newblock The devil is in the margin: Margin-based label smoothing for network calibration.
\newblock In \emph{Proceedings of the IEEE/CVF Conference on Computer Vision and Pattern Recognition}, pages 80--88, 2022.

\bibitem[Moor et~al.(2023)Moor, Banerjee, Abad, Krumholz, Leskovec, Topol, and Rajpurkar]{moor2023foundation}
Michael Moor, Oishi Banerjee, Zahra Shakeri~Hossein Abad, Harlan~M Krumholz, Jure Leskovec, Eric~J Topol, and Pranav Rajpurkar.
\newblock Foundation models for generalist medical artificial intelligence.
\newblock \emph{Nature}, 616\penalty0 (7956):\penalty0 259--265, 2023.

\bibitem[Murugesan et~al.(2024)Murugesan, Silva-Rodr{\'\i}guez, Ayed, and Dolz]{murugesan2024robust}
Balamurali Murugesan, Julio Silva-Rodr{\'\i}guez, Ismail~Ben Ayed, and Jose Dolz.
\newblock Robust calibration of large vision-language adapters.
\newblock In \emph{European Conference on Computer Vision}, pages 147--165. Springer, 2024.

\bibitem[Naeini et~al.(2015)Naeini, Cooper, and Hauskrecht]{ECE}
Mahdi~Pakdaman Naeini, Gregory Cooper, and Milos Hauskrecht.
\newblock Obtaining well calibrated probabilities using bayesian binning.
\newblock In \emph{Proceedings of the AAAI conference on artificial intelligence}, volume~29, 2015.

\bibitem[Niculescu-Mizil and Caruana(2005)]{niculescu2005predicting}
Alexandru Niculescu-Mizil and Rich Caruana.
\newblock Predicting good probabilities with supervised learning.
\newblock In \emph{Proceedings of the 22nd international conference on Machine learning}, pages 625--632, 2005.

\bibitem[Nixon et~al.(2019)Nixon, Dusenberry, Zhang, Jerfel, and Tran]{ACE}
Jeremy Nixon, Michael~W Dusenberry, Linchuan Zhang, Ghassen Jerfel, and Dustin Tran.
\newblock Measuring calibration in deep learning.
\newblock In \emph{CVPR workshops}, volume~2, 2019.

\bibitem[Nori et~al.(2023)Nori, Lee, Zhang, Carignan, Edgar, Fusi, King, Larson, Li, Liu, et~al.]{nori2023can}
Harsha Nori, Yin~Tat Lee, Sheng Zhang, Dean Carignan, Richard Edgar, Nicolo Fusi, Nicholas King, Jonathan Larson, Yuanzhi Li, Weishung Liu, et~al.
\newblock Can generalist foundation models outcompete special-purpose tuning? case study in medicine.
\newblock \emph{arXiv preprint arXiv:2311.16452}, 2023.

\bibitem[Ovadia et~al.(2019)Ovadia, Fertig, Ren, Nado, Sculley, Nowozin, Dillon, Lakshminarayanan, and Snoek]{ovadia2019can}
Yaniv Ovadia, Emily Fertig, Jie Ren, Zachary Nado, David Sculley, Sebastian Nowozin, Joshua Dillon, Balaji Lakshminarayanan, and Jasper Snoek.
\newblock Can you trust your model's uncertainty? evaluating predictive uncertainty under dataset shift.
\newblock \emph{Advances in neural information processing systems}, 32, 2019.

\bibitem[Platt et~al.(1999)]{platt1999probabilistic}
John Platt et~al.
\newblock Probabilistic outputs for support vector machines and comparisons to regularized likelihood methods.
\newblock \emph{Advances in large margin classifiers}, 10\penalty0 (3):\penalty0 61--74, 1999.

\bibitem[Popordanoska et~al.(2022)Popordanoska, Sayer, and Blaschko]{ECE-KDE}
Teodora Popordanoska, Raphael Sayer, and Matthew Blaschko.
\newblock A consistent and differentiable lp canonical calibration error estimator.
\newblock \emph{Advances in Neural Information Processing Systems}, 35:\penalty0 7933--7946, 2022.

\bibitem[Rahman et~al.(2021)Rahman, Khandakar, Qiblawey, Tahir, Kiranyaz, Kashem, Islam, Al~Maadeed, Zughaier, Khan, et~al.]{COVID}
Tawsifur Rahman, Amith Khandakar, Yazan Qiblawey, Anas Tahir, Serkan Kiranyaz, Saad Bin~Abul Kashem, Mohammad~Tariqul Islam, Somaya Al~Maadeed, Susu~M Zughaier, Muhammad~Salman Khan, et~al.
\newblock Exploring the effect of image enhancement techniques on covid-19 detection using chest x-ray images.
\newblock \emph{Computers in biology and medicine}, 132:\penalty0 104319, 2021.

\bibitem[Sharifdeen et~al.(2025)Sharifdeen, Munir, Baliah, Khan, and Khan]{sharifdeen2025tpt}
Ashshak Sharifdeen, Muhammad~Akhtar Munir, Sanoojan Baliah, Salman Khan, and Muhammad~Haris Khan.
\newblock O-tpt: Orthogonality constraints for calibrating test-time prompt tuning in vision-language models.
\newblock \emph{arXiv preprint arXiv:2503.12096}, 2025.

\bibitem[Stein et~al.(2018)Stein, Wu, Carr, Shih, Dulkowski, Kalpathy-Cramer, et~al.]{RSNA18}
Anouk Stein, Carol Wu, Chris Carr, George Shih, Jamie Dulkowski, J~Kalpathy-Cramer, et~al.
\newblock Rsna pneumonia detection challenge.
\newblock \emph{Mountain View: Kaggle}, 2018.

\bibitem[Wang et~al.(2024)Wang, Wang, Wang, Zhang, Zhou, and Wei]{wang2024open}
Shuoyuan Wang, Jindong Wang, Guoqing Wang, Bob Zhang, Kaiyang Zhou, and Hongxin Wei.
\newblock Open-vocabulary calibration for fine-tuned clip.
\newblock In \emph{International Conference on Machine Learning}, pages 51734--51754. PMLR, 2024.

\bibitem[Wang et~al.(2022)Wang, Wu, Agarwal, and Sun]{MEDCLIP}
Zifeng Wang, Zhenbang Wu, Dinesh Agarwal, and Jimeng Sun.
\newblock Medclip: Contrastive learning from unpaired medical images and text.
\newblock In \emph{Proceedings of the Conference on Empirical Methods in Natural Language Processing. Conference on Empirical Methods in Natural Language Processing}, volume 2022, page 3876, 2022.

\bibitem[Wei et~al.(2022)Wei, Xie, Cheng, Feng, An, and Li]{wei2022mitigating}
Hongxin Wei, Renchunzi Xie, Hao Cheng, Lei Feng, Bo~An, and Yixuan Li.
\newblock Mitigating neural network overconfidence with logit normalization.
\newblock In \emph{International conference on machine learning}, pages 23631--23644. PMLR, 2022.

\bibitem[Yoon et~al.(2024)Yoon, Yoon, Tee, Hasegawa-Johnson, Li, and Yoo]{yoon2024c}
Hee~Suk Yoon, Eunseop Yoon, Joshua Tian~Jin Tee, Mark Hasegawa-Johnson, Yingzhen Li, and Chang~D Yoo.
\newblock C-tpt: Calibrated test-time prompt tuning for vision-language models via text feature dispersion.
\newblock \emph{arXiv preprint arXiv:2403.14119}, 2024.

\bibitem[Zhang et~al.(2023)Zhang, Xu, Usuyama, Xu, Bagga, Tinn, Preston, Rao, Wei, Valluri, et~al.]{BIOMEDCLIP}
Sheng Zhang, Yanbo Xu, Naoto Usuyama, Hanwen Xu, Jaspreet Bagga, Robert Tinn, Sam Preston, Rajesh Rao, Mu~Wei, Naveen Valluri, et~al.
\newblock Biomedclip: a multimodal biomedical foundation model pretrained from fifteen million scientific image-text pairs.
\newblock \emph{arXiv preprint arXiv:2303.00915}, 2023.

\bibitem[Zhao et~al.(2023)Zhao, Liu, Wu, Wang, Li, Wang, Teng, Liu, Cui, Wang, et~al.]{zhao2023clip}
Zihao Zhao, Yuxiao Liu, Han Wu, Mei Wang, Yonghao Li, Sheng Wang, Lin Teng, Disheng Liu, Zhiming Cui, Qian Wang, et~al.
\newblock Clip in medical imaging: A comprehensive survey.
\newblock \emph{arXiv preprint arXiv:2312.07353}, 2023.

\bibitem[Zhou et~al.(2022)Zhou, Yang, Loy, and Liu]{zhou2022learning}
Kaiyang Zhou, Jingkang Yang, Chen~Change Loy, and Ziwei Liu.
\newblock Learning to prompt for vision-language models.
\newblock \emph{International Journal of Computer Vision}, 130\penalty0 (9):\penalty0 2337--2348, 2022.

\end{thebibliography}
